\pdfoutput=1

\documentclass[11pt]{article}

\usepackage[]{acl}

\usepackage{times}
\usepackage{latexsym}
\usepackage{multicol}
\usepackage{multirow}
\usepackage{graphicx}
\usepackage[T1]{fontenc}

\usepackage[utf8]{inputenc}

\usepackage{microtype}

\usepackage{inconsolata}

\title{FinGen: A Dataset for Argument Generation in Finance}

\author{Chung-Chi Chen,\textsuperscript{1} Hiroya Takamura,\textsuperscript{1} Ichiro Kobayashi,\textsuperscript{2}  Yusuke Miyao\textsuperscript{3}
\\
 \textsuperscript{1} Artificial Intelligence Research Center, AIST, Japan \\
\textsuperscript{2} Ochanomizu University, Japan \\
 \textsuperscript{3} University of Tokyo, Japan \\
   \texttt{c.c.chen@acm.org, takamura.hiroya@aist.go.jp,}\\ \texttt{koba@is.ocha.ac.jp, yusuke@is.s.u-tokyo.ac.jp}\\
}

\begin{document}
\maketitle
\begin{abstract}
Thinking about the future is one of the important activities that people do in daily life. Futurists also pay a lot of effort into figuring out possible scenarios for the future. We argue that the exploration of this direction is still in an early stage in the NLP research. To this end, we propose three argument generation tasks in the financial application scenario. Our experimental results show these tasks are still big challenges for representative generation models. Based on our empirical results, we further point out several unresolved issues and challenges in this research direction.
\end{abstract}

\section{Introduction}
Argument generation is a critical issue in argument mining research and one of the most recently focused topics. 
It can be separated into three steps~\cite{hua-etal-2019-argument-generation}: (1) evidence retrieval~\cite{hua-wang-2018-neural}, (2) structure/strategy planning~\cite{saint-dizier-2016-challenges,wachsmuth-etal-2018-argumentation,li-etal-2022-neural,wambsganss-niklaus-2022-modeling}, and (3) argument synthesis~\cite{sato-etal-2015-end,el-baff-etal-2019-computational,schiller-etal-2021-aspect,holtermann-etal-2022-fair,alshomary-etal-2022-moral}.
Based on this process, we can conclude that recent research focuses on conditionally summarizing arguments, which generates the argument based on a specific stance and strategy with given evidence. 
That is, recent studies have focused on summarizing past and current evidence and forming a persuasive expression by selecting appropriate structures and strategies.
We contend that generating arguments with a forward-looking, future-oriented perspective has received little attention in previous research.
Thus, we present three forward-looking argument generation tasks to discuss what is solved and what remains challenging.

At this moment, nobody knows what will happen in the future, but people try to make claims based on the available evidence in the past and the plausible scenarios for the future~\cite{amer2013review,mori2020image}.
We notice that the discussions about the future frequently happen in the managers' and investors' narratives, because managers need to explain the future operation and possible impact to investors, and investors are used to imagining the possible scenarios and link them with asset values.
We think that both managers' expectations and investors' imagination about the future are kinds of forward-looking argument, and is considered as investment suggestions in this paper. 
Therefore, in this study, we propose three tasks for learning to generate forward-looking arguments as managers and investors as follows: 
\begin{itemize}
    \small
    \item \textbf{Evidence2Claim}: 
    Given the evidence, which is the premises of the claims, generate forward-looking claims on the company's future operation, i.e., subjective opinions about the future operation.
    
    \item \textbf{Chart2Argument}: Given a price chart with investors' notes, generate a summary of past price movements and investors' investment suggestions.
    
    \item \textbf{News2Argument}: Given a news article, generate possible scenarios and investment suggestions.
\end{itemize}
These tasks are happening in the daily financial market, and both investors and managers are still finding a way to figure out possible futures. 
Table~\ref{fig:Example} provides an example of each task, and we detail the proposed tasks and datasets in Section~\ref{sec:dataset}.

\begin{table*}[t]
  \centering
  \includegraphics[width=16cm]{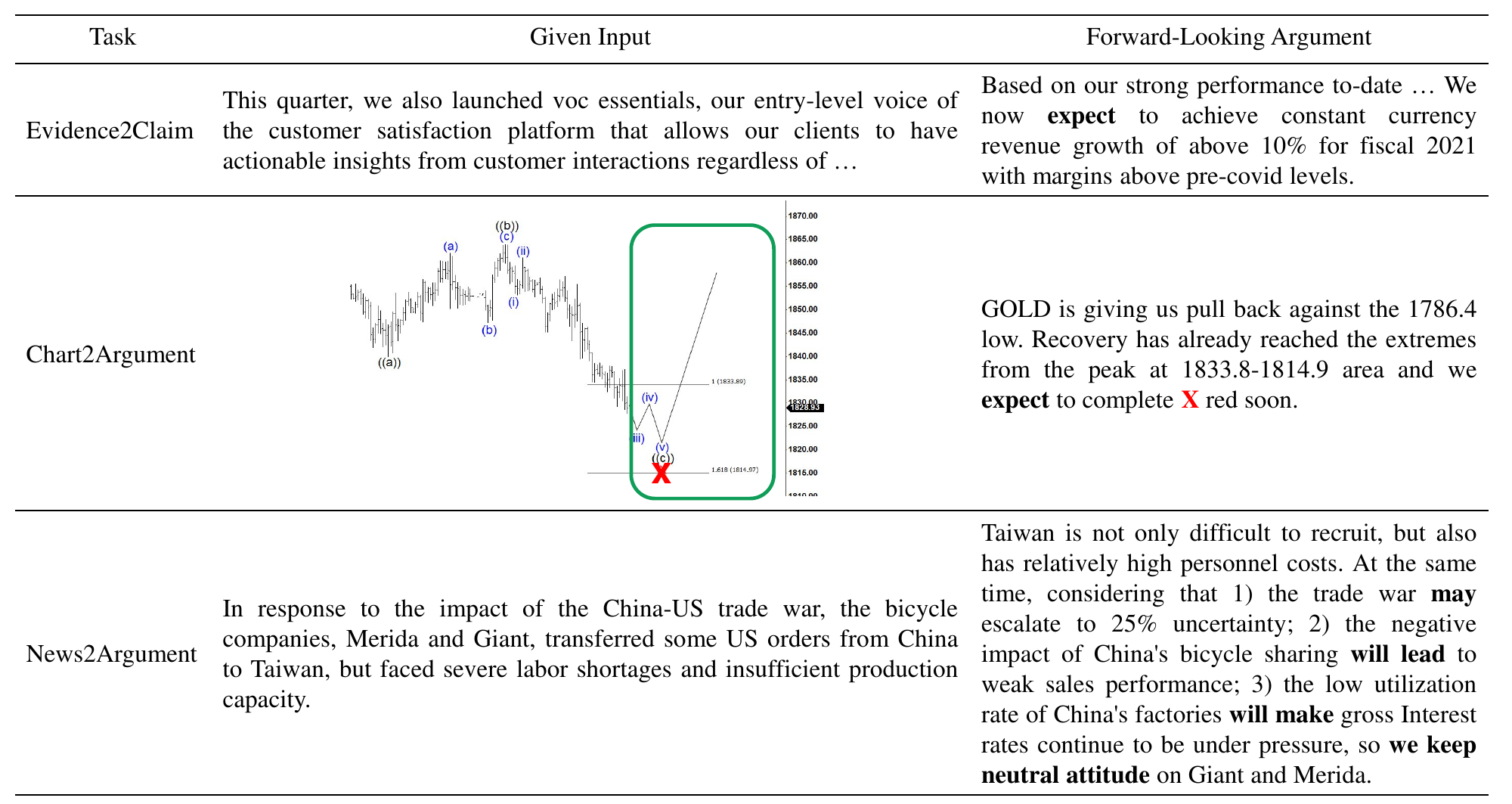}
  \caption{Examples of inputs for the proposed tasks, accompanied by forward-looking arguments.}
  \label{fig:Example}
\end{table*}

\begin{table*}[t]
  \centering
  \resizebox{\textwidth}{!}{
    \begin{tabular}{lcccccr}
     \hline
    Task & Language & Input & Aspect & Expresser & Source & \multicolumn{1}{c}{Instances} \\
    \hline
    Evidence2Claim (E2C) & English & Text  & Fundamental Analysis & Manager & Earnings Conference Call & 34,933 \\
    Chart2Argument (C2A) & English & Chart & Technical Analysis & Investor & Blog  & 6,987 \\
    News2Argument (N2A) & Chinese & Text  & Fundamental Analysis & Analyst & Professional Report & 2,004 \\
    \hline
    \end{tabular}%
    }
  \caption{Overview of FinGen.}
  \label{tab:Overview of datasets}%
\end{table*}%

As discussed in previous studies~\cite{narayan2021planning,narayan-etal-2022-well}, entities play an important role in text generation. 
In financial narratives, there are two important parts that play the same role as entities: financial terms and numerals.
Following in this vein, we propose a keyterm-guided approach for improving performance in the Evidence2Claim task and investigate various image-text encoder-decoder architectures for the Chart2Argument task. 
We further analyze the results from several aspects to discuss the challenge of these tasks.

\section{Tasks and Datasets}
\label{sec:dataset}
This section provides the design of the proposed three tasks and the details of the proposed dataset, FinGen.
Some examples are listed in Table~\ref{fig:Example}, and an overview of all the datasets is given in
Table~\ref{tab:Overview of datasets}.
These datasets are written in either English or Chinese and contain two kinds of input data (text and chart). 
Because different expressers focus on different viewpoints, the proposed tasks include the exploration of fundamental analysis (based on financial statement analysis and companies' operations) and technical analysis (based on the historical price chart).
The sources are also diverse, including talk transcriptions (earnings conference call), formal reports, and social media posts (blog).
For all datasets, we use 80\% of instances for training and the remaining 20\% for evaluation.

\subsection{Evidence2Claim}
As the example shown in Table~\ref{fig:Example}, companies' managers always state what happened and then make a claim related to monetary terms based on the stated evidence.
Thus, in the Evidence2Claim task, we aim to explore to what extent models can generate forward-looking claims as managers.

We collected the transcripts of the earnings conference call from SeekingAlpha,\footnote{\url{https://seekingalpha.com/}} and separated each paragraph into evidence and forward-looking claims. 
Because forward-looking claims have been discussed in financial literature for a long time~\cite{li2010information,wang2013voluntary,muslu2015forward,bozanic2018management,cazier2020firms},  previous research has shown that keyword-based methods can achieve high agreement with human annotations. 
\citet{alhamzeh2022passau21} also indicate that some words like ``expect'' related to forward-looking statements play an important role in identifying forward-looking claims. 
Inspired by these previous studies, we adopt the following keywords for identifying forward-looking claims in the earnings conference calls: \textit{expect, anticipate, estimate, intend, will, could, future, believe, predict, and would}.

After the identification, we take the rest of the sentences in the same paragraph as the evidence, i.e., the model's input. 
After this process, we get 34,933 instances for the Evidence2Claim task.

\subsection{Chart2Argument}
Technical analysis is widely used by investors for stock movement prediction. 
As shown in Table~\ref{fig:Example}, investors will note a few turning points on the historical price chart and try to imagine the future price movement (as the green framed part). 
After that, they will write down their analysis and forward-looking claims. 
In the Chart2Argument task, we probe the ability of the representative encoder-decoder architectures to generate arguments based on the investors' noted price chart. 

We collect the dataset from Elliott Wave Forecast Blog.\footnote{\url{https://elliottwave-forecast.com/}}
All posts on this blog analyze the price chart based on the same technical analysis method, Elliott Wave Theory.
Elliott Wave Theory focuses on the patterns of historical prices and makes forecasts for the price movement based on the corrective price waves.
Finally, we obtained 6,987 chart-argument pairs for the Blog.

\subsection{News2Argument}
Foreseeing the possible scenarios and making claims based on these scenarios are essential tasks of professional analysts. 
As an example in Table~\ref{fig:Example}, analysts first provide three possible scenarios, and then express their opinion (neutral attitude). 
This task is different from Evidence2Claim and Chart2Argument, which summarize the given input with forward-looking claims. 
In the News2Argument tasks, models need to (1) perform scenario planning, i.e., generate possible future events, and (2) make forward-looking claims based on the generated scenarios instead of the given news articles.

We collect 2,004 Chinese news-argument pairs from the professional analysts' reports published by securities company from 2018/11 to 2020/01.
All arguments are written by professional analysts and sent to customers every trading day. 
Note that analysts will not write comments on all news articles. 
They only select a few important daily news items for making comments. 
That is the reason why we can only get a few instances even though we have collected this dataset for over one year.

\section{Methods}
\subsection{Pretrained Models}
For Evidence2Claim and News2Argument tasks, we explore with T5, multilingual T5 (mT5)~\cite{raffel2020exploring}, BART~\cite{lewis2020bart}, Pegasus~\cite{zhang2020pegasus}, and Pegasus-Chinese provided by UER toolkit~\cite{zhao2019uer}.

For the Chart2Argument task, we use the transformer-based encoder-decoder architecture, TrOCR~\cite{li2021trocr}, which uses BEiT~\cite{bao2021beit} as an encoder and RoBERTa~\cite{liu2019roberta} as a decoder.
We further explore several representative image encoders, including DeiT~\cite{touvron2021training}, Swin~\cite{liu2021swin}, and ViT~\cite{dosovitskiy2021image}.
We use BERT~\cite{devlin-etal-2019-bert}, RoBERTa~\cite{liu2019roberta}, and LinkBERT~\cite{yasunaga-etal-2022-linkbert} as decoders.

\subsection{Keyterm-Guided Method}
Previous studies showed that entities play an important role in text generation~\cite{narayan2021planning,narayan-etal-2022-well}, and numerals provide crucial information in financial documents~\cite{chen2021opinion}. 
Inspired by previous studies, we propose a keyterm-guided method for emphasizing two kinds of crucial components in financial narratives: financial terms and numerals. 
We reframe the input text with two steps:
(1) \textbf{Emphasize Numeral (EmphNum)}: We use a tag ([NUM]) to highlight the numerals in the input text. 
(2) \textbf{Repeat Financial Term (RFT)}: We repeat the financial terms at the end of the input text, and use ``|'' symbol to separate financial terms.
To identify financial terms, we performed the longest match with the domain-specific dictionary, Investopedia Dictionary.
For example, the statement ``Resulting from 5\% \textit{organic sales} growth, \$14 million in pricing, and \$36 million of favorable \textit{foreign exchange}.'' will become ``
Resulting from [NUM] 5\% [NUM] organic sales growth, [NUM] \$14 [NUM] million in pricing and [NUM] \$36 [NUM] million of favorable foreign exchange. [SEP] organic sales | foreign exchange'' after reframing.

\section{Experiments}

\subsection{Experimental Results}
\label{sec:Experimental Results}
We adopt ROUGE metrics~\cite{lin-2004-rouge} and BERT Scores~\cite{zhang2019bertscore} for evaluating all results.
Table~\ref{tab:Evidence2Claim} shows the experimental results of the Evidence2Claim task. 
First, we discover that BART excels at this task, and the proposed keyterm-guided method works in all pre-trained models tested. 
Second, we notice that the BERT Score is quite high. 
Based on our observations, we find that models have already learned how to make claims. 
That is, most generated forward-looking claims start with ``We expect'' or ``We believe.''

Table~\ref{tab:Chart2Argument} reports the results of the Chart2Argument task. 
Vit-RoBERTa performs the best among all permutations. 
Additionally, RoBERTa performs the best when using ViT and Swin encoders, and BERT performs the best when using DeiT.
Surprisingly, LinkBERT, which performs well in several benchmarks and contains cross-document knowledge, does not outperform its ancestors.
Table~\ref{tab:News2Argument} shows the results of News2Argument task. 
Because some previous studies~\cite{hua-wang-2020-pair,hu-etal-2022-planet} used news headlines as input to generate opinion articles, we provide the results by using both headlines and news articles as input for comparison.
First, Pegasus performs better than mT5 regardless of which input we use. 
Second, we show the performance gap between using headlines and news articles.

\begin{table}[t]
  \centering
  \resizebox{\columnwidth}{!}{
    \begin{tabular}{lrrrr}
    \hline
    & \multicolumn{1}{c}{ROUGE-1} & \multicolumn{1}{c}{ROUGE-2} & \multicolumn{1}{c}{ROUGE-L} & \multicolumn{1}{c}{BERT Score} \\
    \hline
    T5    & 0.2476 & 0.0590 & 0.1866 & 0.8660 \\
    + Keyterm-Guided & 0.2518 & 0.0619 & \textbf{0.1901} & 0.8637 \\
    \hline
    BART  & 0.2516 & 0.0613 & 0.1892 & \textbf{0.8682} \\
    + Keyterm-Guided & \textbf{0.2526} & \textbf{0.0649} & 0.1895 & 0.8648 \\
    \hline
    Pegasus & 0.2395 & 0.0623 & 0.1838 & 0.8612 \\
    + Keyterm-Guided & 0.2418 & 0.0637 & 0.1851 & 0.8621 \\
    \hline
    \end{tabular}%
    }
  \caption{Results of Evidence2Claim task.}
  \label{tab:Evidence2Claim}%
\end{table}%

\begin{table}[t]
  \centering
  \resizebox{\columnwidth}{!}{
    \begin{tabular}{llrrrr}
    \hline
    \multicolumn{1}{c}{Encoder} & \multicolumn{1}{c}{Decoder} & \multicolumn{1}{c}{ROUGE-1} & \multicolumn{1}{c}{ROUGE-2} & \multicolumn{1}{c}{ROUGE-L} & \multicolumn{1}{c}{BERT Score} \\
    \hline
    \multicolumn{2}{l}{TrOCR} & 0.2167 & 0.0958 & 0.1517 & 0.81015 \\
    \hline
    \multirow{3}[2]{*}{DeiT} & BERT  & 0.2410 & 0.1120 & 0.1646 & 0.82776 \\
          & RoBERTa & 0.2333 & 0.1083 & 0.1615 & 0.82536 \\
          & LinkBERT & 0.2373 & 0.1118 & 0.1604 & 0.83267 \\
    \hline
    \multirow{3}[2]{*}{Swin} & BERT  & 0.2323 & 0.1071 & 0.1588 & 0.82660 \\
          & RoBERTa & 0.2440 & 0.1198 & 0.1685 & 0.83143 \\
          & LinkBERT & 0.2431 & 0.1170 & 0.1665 & \textbf{0.83514} \\
    \hline
    \multirow{3}[2]{*}{ViT} & BERT  & 0.2336 & 0.1019 & 0.1554 & 0.82535 \\
          & RoBERTa & \textbf{0.2444} & \textbf{0.1213} & \textbf{0.1695} & 0.83311 \\
          & LinkBERT & 0.2361 & 0.1105 & 0.1598 & 0.83283 \\
    \hline
    \end{tabular}%
    }
  \caption{Results of Chart2Argument task.}
  \label{tab:Chart2Argument}%
\end{table}%

\begin{table}[t]
  \centering
  \resizebox{\columnwidth}{!}{
    \begin{tabular}{lrrrr}
    \hline
          & \multicolumn{1}{c}{ROUGE-1} & \multicolumn{1}{c}{ROUGE-2} & \multicolumn{1}{c}{ROUGE-L} & \multicolumn{1}{c}{BERT Score} \\
    \hline
    mT5 - Headline & 0.1238 & 0.0475 & 0.1195 & 0.6471 \\
    mT5 - Article & 0.1560 & 0.0629 & 0.1499 & 0.6648 \\
    \hline
    Pegasus - Headline & 0.2098 & 0.0951 & 0.1997 & 0.7034 \\
    Pegasus - Article & \textbf{0.2362} & \textbf{0.1083} & \textbf{0.2273} & \textbf{0.7128} \\

    \hline
    \end{tabular}%
    }
  \caption{Results of News2Argument task.}
  \label{tab:News2Argument}%
\end{table}%

\begin{table}[t]
  \centering
  \resizebox{\columnwidth}{!}{
    \begin{tabular}{lrr}
    \hline
          & \multicolumn{1}{c}{Appear in Input } & \multicolumn{1}{c}{Not Appear in Input} \\
    \hline
    T5    & 52.54\% & 9.30\% \\
    + Keyterm-Guided & \textbf{53.34\%} & 9.22\% \\
    \hline
    BART  & 49.11\% & \textbf{12.44\%} \\
    + Keyterm-Guided & 49.43\% & 11.15\% \\
    \hline
    Pegasus & 48.53\% & 11.38\% \\
    + Keyterm-Guided & 48.72\% & 11.54\% \\
    \hline
    \end{tabular}%
    }
  \caption{Analysis of financial terms. (Not) Appear in Input indicates that (none of) at least one financial terms in the ground truth also appears in the input. Percentage of at least one financial term appears in both ground truth and the generated text.}
  \label{tab:Analysis of financial terms}%
\end{table}%

\begin{table}[t]
  \centering
  \resizebox{\columnwidth}{!}{
    \begin{tabular}{lrr}
    \hline
          & \multicolumn{1}{c}{W/ Number} & \multicolumn{1}{c}{Numeral Hallucination} \\
    \hline
    Answer & 49.35\% & \multicolumn{1}{r}{-} \\
    \hline
    T5    & 25.63\% & 17.12\% \\
    + Keyterm-Guided & 23.93\% & 14.98\% \\
    \hline
    BART  & 32.32\% & 20.99\% \\
    + Keyterm-Guided & 32.13\% & 19.38\% \\
    \hline
    Pegasus & 32.80\% & 21.84\% \\
    + Keyterm-Guided & 33.14\% & 21.05\% \\
    \hline
    \end{tabular}%
    }
  \caption{Analysis of numerals.}
  \label{tab:Analysis of numerals}%
\end{table}%

\begin{table*}[t]
  \centering
  \small
  \resizebox{\textwidth}{!}{
    \begin{tabular}{clp{34em}}
    \hline
    Index &       & \multicolumn{1}{c}{} \\
    \hline
    \multirow{16}[0]{*}{E2C-1} & \multirow{8}[0]{*}{Given Input} & These are all projects and milestones where we have strong visibility on the resources needed to complete them, both human and financial. Importantly, we have cash ready on hand with nearly \$200 million in the bank. We also have the expertise needed to make these milestones happen, with strong business foundations, scalable systems, operational controls in place and senior leadership and human capital over 1,500 people strong, 10 of whom are specifically dedicated swat team focused on the integration, project management and best practices and execution of the milestones that drive our 2022 projections. \\
    \cline{2-3}
          & \multirow{3}[0]{*}{Ground Truth} & As we continue to scale operations in our 7 states, you will see a significant presence in states that matter.In 2022, 6 of our 7 states will contribute over \$100 million of revenue to our \$725 million revenue target. \\
    \cline{2-3}
          & \multirow{2}[0]{*}{T5}   & We believe that our strong financial position will enable us to continue to grow our business and deliver on our long-term goals. \\
    \cline{2-3}
          & \multirow{2}[0]{*}{BART}  & We will continue to work with our customers and partners to ensure that we are able to complete these milestones in a timely manner. \\
    \cline{2-3}
          & Pegasus & We believe we are well-positioned to execute on our 2021 and 2022 growth plans. \\
    \hline
    \multirow{9}[0]{*}{E2C-2} & \multirow{2}[0]{*}{Given Input} & However, we recorded our first \textbf{net loss} ever affected by, non-recurring charges for more than €300 million. \\
    \cline{2-3}
          & \multirow{3}[0]{*}{Ground Truth} & Given the current situation we will recommend to the next AGM, not to pay dividend in 2020. And you can count on us, to continue our ongoing efforts to adjust our cost base, our working capital and capex, on revenue level. \\
    \cline{2-3}
          & T5    & We expect to see a \textcolor{red}{net loss} of €100 million in the second half of 2021. \\
    \cline{2-3}
          & BART  & We expect the \textcolor{red}{non-recurring charges} to continue in the second half of the year. \\
    \cline{2-3}
          & \multirow{2}[0]{*}{Pegasus} & For the full year, we expect \textcolor{red}{adjusted EBITDA} to be in the range of 1.2 billion to 1.3 billion. \\
    \hline
    \multirow{11}[0]{*}{E2C-3} & \multirow{2}[0]{*}{Given Input} & \textbf{Corus} represents a compelling investment with a multiple of under 6 times, a free cash flow yield of almost 25\% and an attractive dividend yield of 4\%. \\
    \cline{2-3}
          & \multirow{3}[0]{*}{Ground Truth} & Our shareholders will benefit both from the reopening of the economy, our disciplined capital allocation policy and our commitment to deliver consolidated revenue growth in the years ahead. \\
    \cline{2-3}
          & \multirow{2}[0]{*}{T5}   & We believe that \textcolor{red}{Corus} is a strong investment opportunity with a strong balance sheet and a strong balance sheet. \\
    \cline{2-3}
          & \multirow{2}[0]{*}{BART}  & We continue to believe that \textcolor{red}{Corus} is well positioned to benefit from the secular tailwinds that we are seeing in our business \\
    \cline{2-3}
          & \multirow{2}[0]{*}{Pegasus} & We continue to believe that \textcolor{red}{Corus} is well positioned to deliver long-term value to our shareholders. \\
    \hline
    \end{tabular}%
    }
  \caption{Cases studies of Evidence2Claim.}
  \label{tab:Cases Studies of Evidence2Claim}%
\end{table*}%

\begin{figure*}[t]
  \centering
  \includegraphics[width=11cm]{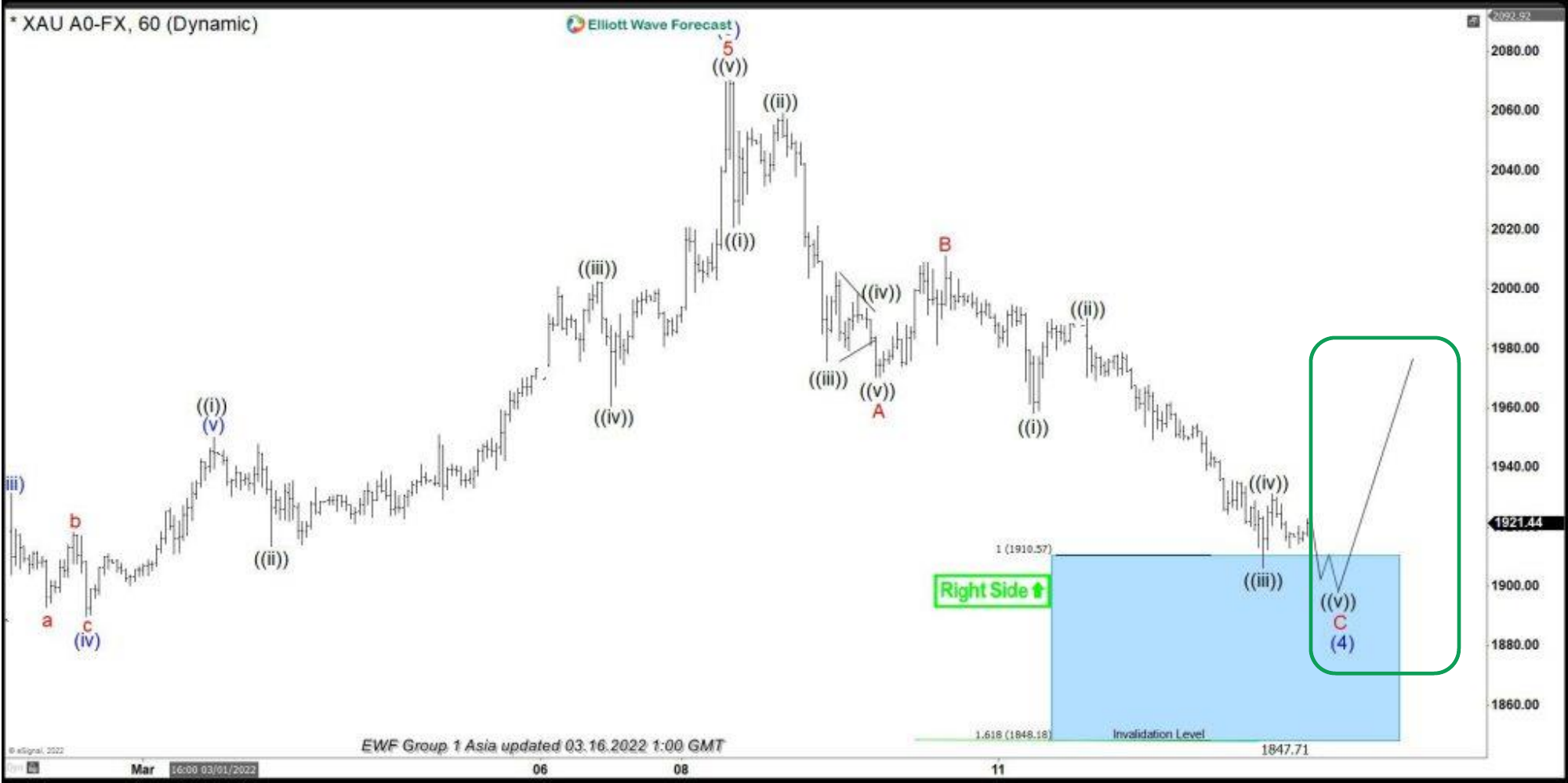}
  \caption{Example input of Chart2Argument task.}
  \label{fig:Example input of Chart2Argument task}
\end{figure*}

\subsection{Challenges}
\label{challenge}

Based on our observation, we notice two important challenges in these tasks. 
Firstly, models perform well in copying the financial terms from the input, but worse when the financial terms do not appear in the input. 
Table~\ref{tab:Analysis of financial terms} provides the evaluation of this phenomenon. 
These results also show that the proposed keyterm-guided method helps all models in the copying case.
Secondly, models generate fewer numerals than they should. 
Table~\ref{tab:Analysis of numerals} shows the statistics of numerals. 
There are about 50\% answers that contain at least one numeral, but only 23\% to 33\% of generated results contain at least one numeral. 
Additionally, we also find that models generate numerals when the answer does not contain any numeral.
We call it: \textit{numeral hallucination}.
Table~\ref{tab:Analysis of numerals} shows the ratios of numeral hallucination of each model.
We find that the proposed method can reduce this hallucination regardless of the language model used.

\section{Case Studies}
\label{sec:Examples and Case Studies}
Table~\ref{tab:Cases Studies of Evidence2Claim} provides some cases of the Evidence2Claim task. 
Models perform well in generating human-like forward-looking claims, and E2C-1 is an example of how models generate great and plausible forward-looking claims.
E2C-2 shows an example that T5 copies the financial term (net loss) from the given input, and all models generate forward-looking claims on different financial terms. 
It is intuitive that managers seldom make claims on the future ``net loss''. 
One of the interesting future directions is how to let models learn to infer the possible financial term chain.
E2C-3 points out that current representative models seem to simply change a way to summarize the given input instead of making claims with causality inference.

\begin{table*}[t]
  \small
  \centering
    \begin{tabular}{lp{18em}p{24em}}
    \hline
    Index & \multicolumn{1}{c}{Ground Truth} & \multicolumn{1}{c}{Generated Argument} \\
    \hline
    \multirow{6}[0]{*}{C2A-1}    & \textcolor{red}{\$SAN} : Sanofi Shows Monthly Bullish Structure May 6, 2020 By EWF Helgi Sanofi S.A. is a French multinational pharmaceutical company being within 10 largest pharmaceutical companies in the world. & \textcolor{red}{\$FP} : Pattern of Total stock can be better January 19, 2020 By EWF Helgi Total SA is a German multinational company which designs, develops, and markets smart and connected premium electric vehicles, driving innovations in next generation technologies in connectivity, autonomous driving, and artificial intelligence.  \\
    \hline
    \multirow{5}[0]{*}{C2A-2}   & Elliott Wave View: Apple in Zigzag Correction March 26, 2020 By EWFHendra Short Term Elliott Wave view in Apple (AAPL) suggests that cycle from \textcolor{red}{February 13, 2020} high has ended at 212.6 low as wave w. & Elliott Wave View: Apple (AAPL) Continues to Benefit from Covid-19 Pandemic September 4, 2020 By EWFHendra Short term Elliott Wave view in Apple (Ticker: AAPL) suggests the rally from \textcolor{red}{August 15, 2020} low is unfolding as a 5 waves impulse Elliott Wave structure. \\
    \hline
    C2A-3    & \multicolumn{1}{l}{wave ((i)) ended at \textcolor{red}{88.20}} & \multicolumn{1}{l}{wave ((i)) ended at \textcolor{red}{64.49}} \\
    \hline
    \multirow{4}[0]{*}{C2A-4}    & Thus \textcolor{red}{provided buying opportunity to our members} at the blue box area looking for the next extension higher or for 3 wave reaction higher at least. & We suggested members to \textcolor{red}{keep on selling} the rallies in the blue box areas when get 3 waves down,while favoring buying the short side.  \\
    \hline
    \end{tabular}%

  \caption{Cases studies of Chart2Argument with ViT-RoBERTa's results.}
  \label{tab:Cases Studies of Chart2Argument}%
\end{table*}%

Table~\ref{tab:Cases Studies of Chart2Argument} provides examples of the output of the best-performing model, ViT-RoBERTa, in the Chart2Argumet task. 
We notice that most generated arguments are fluent. 
However, a close examination of the generated results and the provided price chart reveals some issues.
Figure~\ref{fig:Example input of Chart2Argument task} shows an example input of the Chart2Argument task. 
Some parts of the chart are crucial for the quality of the generated arguments.
First, the asset name will be discussed in the chart's top-left corner (XAU, which represents the gold price). 
We notice that models perform well in generating the right discussion target if the target has appeared in the training set. 
The C2A-1 and C2A-2 in Table~\ref{tab:Cases Studies of Chart2Argument}, where the generated arguments are based on the best-performing model (ViT-RoBERTa), provide examples for selecting wrong (\$SAN and \$FP in C2A-1) and correct (Apple in C2A-2) targets, respectively. 
Second, C2A-2 indicates the other challenge of the proposed task: identifying the crucial point(s) in the past price chart.
In C2A-2, the model generates a time stamp that is different from the ground truth. 
That is, because there have been many turning points in history, determining which are significant and worth discussing based on the technical analysis method we used is a critical task.
Third, the chart's y-axis (right side of the chart) provides information on the price level. 
Models need to identify such numerical details for generating the results as the ground truth of C2A-3. 
Or, the numeral information will be incorrect as the generated argument of C2A-3.
Fourth, the right-hand side of the chart (green framed part)\footnote{The green frame is added by the authors for the explanation.} is the expected price movement that is drawn by the investor, which contains the investor's forward-looking view of the price movement.
However, it is also a challenge for models to understand this part. 
C2A-4 in Table~\ref{tab:Cases Studies of Chart2Argument} shows an example that the model generates a forward-looking claim with a sentiment that is different from the ground truth. 

Table~\ref{tab:Cases Studies of News2Argument} provides examples for the News2Argument task. 
As the results of Evidence2Claim and Chart2Argument tasks, most generated results are fluent and plausible.
Based on N2A-1, we find that models perform well if a similar topic has been discussed in the training set. 
However, some fine-grained information may be incorrect.
For example, as shown in N2A-2, the date and the monetary terms are incorrect and greatly differ from those in the ground truth.
Additionally, the degrees of sentiment in the generated arguments in N2A-3 are also different from the ground truth. 
These results show that there is still room to improve the performance of generating expert-like arguments.
We also find that models tend to summarize the given news with the tone of a forward-looking claim, but few perform scenario planning as the instance in Table~\ref{fig:Example}.

\begin{table*}[t]
  \centering
  \small
    \begin{tabular}{llp{38em}}
    \hline
    \multicolumn{1}{c}{Index} &       & \multicolumn{1}{l}{} \\
    \hline
    \multirow{16}[0]{*}{N2A-1} & \multirow{6}[0]{*}{Ground Truth} & According to the survey, LandMark Optoelectronics' 4Q18 revenue will have low double-digit quarterly growth, and the view remains unchanged. The revenue growth momentum came from the silicon light plan and the significant growth rate of the revenue of products above 10g in the quarter. However, we still have doubts about the future growth rate of silicon photonics products, mainly because the traditional optical communication solutions are mature and cannot be easily replaced in the long-distance market. We maintain a neutral view of LandMark. \\
    \cline{2-3}
          & \multirow{5}[0]{*}{Pegasus-Title} & We believe that LandMark's fourth-quarter revenue is in line with the previous quarterly growth forecast announced by the company, mainly benefiting from the strong demand from the top two customers of the Silicon Light project. However, we recommend investors wait and see for the high price-to-earnings ratio at this stage to evaluate technology stocks. We will make another recommendation when the China-US trade situation becomes more apparent. \\
    \cline{2-3}
          & \multirow{5}[0]{*}{Pegasus-News} & We believe that LandMark's revenue is in line with the previous quarterly growth outlook announced by the law, mainly benefiting from the strong demand from the top two customers of the Silicon Light project. However, we recommend investors wait and see for the high price-to-earnings ratio at this stage to evaluate technology stocks. We will make another recommendation when the China-US trade situation becomes more apparent. \\
    \hline
    \multirow{3}[0]{*}{N2A-2} & Ground Truth & \multicolumn{1}{l}{Estimated EPS for 2018/19 is 3.08/3.17 yuan.} \\
          & Pegasus-Title & \multicolumn{1}{l}{Estimated \textcolor[rgb]{ 1,  0,  0}{2019/20} EPS is \textcolor[rgb]{ 1,  0,  0}{10.06/12.79} yuan.} \\
          & Pegasus-News & \multicolumn{1}{l}{Estimated \textcolor[rgb]{ 1,  0,  0}{2019/20} EPS is 3.44/3.68 yuan.} \\
    \hline
    \multirow{4}[0]{*}{N2A-3} & \multirow{2}[0]{*}{Ground Truth} & We believe that a temporary slowdown in growth provides \textcolor[rgb]{ 1,  0,  0}{a good opportunity} for investors to enter the market \\
          & Pegasus-Title & It remains to be seen, and \textcolor[rgb]{ 1,  0,  0}{continued attention} is recommended \\
          & Pegasus-News & \textcolor[rgb]{ 1,  0,  0}{Recommended to follow} \\
    \hline
    \end{tabular}%
    \caption{Cases studies of News2Argument.}
  \label{tab:Cases Studies of News2Argument}%
\end{table*}%

\section{Future Directions}
\label{sec:future direction}
This section discusses where we are now and where we are going.
We separate the discussion into two parts: forward-looking claim generation and scenario planning.
Forward-looking claim generation is a task that focuses on summarizing past events with the forecast. 
On the other hand, scenario planning is a task that has several steps: (1) list all plausible scenarios, (2) select a few scenarios, and (3) generate a forward-looking claim based on the selected scenarios. 
Step (4) may exist in some cases to provide possible operations (solutions) to the selected scenarios. 
Forward-looking claim generation can be considered a kind of extension of forecasting tasks, but scenario planning should not~\cite{amer2013review}. 

\subsection{Forward-Looking Claim Generation}
\label{sec:Forward-Looking Claim Generation}
Claim generation is one of the popular topics in argument mining and argument generation studies. 
It aims to summarize the given evidence based on the given stance with a subjective tone~\cite{hidey-mckeown-2019-fixed,gretz-etal-2020-workweek,atanasova-etal-2020-generating,pan-etal-2021-zero,alshomary-etal-2021-belief,wright-etal-2022-generating}. 
Unlike previous studies, this study focuses on forward-looking claim generation, which should take into account not only available evidence but also future inferences based on causality and event chains.

The BERT Scores reported in Section~\ref{sec:Experimental Results} support that recent models have already been able to generate fluent and plausible sentences regardless of the task and the number of training instances. 
Despite the fact that there are only 1,603 training instances for News2Argument, the BERT Score remains high. 
Additionally, even using the price chart as an input, models can still generate human-like claims.
However, we notice that there is still room to improve the details in the generated claims.
We suggest considering forward-looking claims with the following template, which contains three important parts.
\vspace{1.5mm}

\noindent \textit{We expect the \textbf{[Financial Term]} will be \textbf{[Direction]} for \textbf{[Fine-Grained Estimation]}.
}

\vspace{1.5mm}

\noindent \textit{[Financial Term]} can be sales, revenue, cost, earnings, and any financial terms, and it is the subject of our claims. 
It is important to decide which financial term is related to the evidence, because it is impossible to discuss everything in the given time slot. 
However, we find that models perform well when the target financial term appears in the input but are poor in inferring the term chain (when the target financial terms do not appear in the input). 
Table~\ref{tab:Analysis of financial terms} provides the evaluation of this phenomenon. 
It also shows that the proposed keyterm-guided method helps all models in the copying case.
This leads to the first open issue in forward-looking claim generation: which topic, i.e., \textit{[Financial Term]} in our task, should we consider and discuss?

\textit{[Direction]} can be increase or decrease, and \textit{[Fine-grained Estimation]} is an assessment of the increase/decrease. 
We notice that models may generate ``meet our expectation'' when the professional's opinion is ``lower than our expectation'' (Past Event Hallucination).
Additionally, there may exist Causal Hallucination in the generated summarization. 
For example, models generate ``the decrease of iPhone's sales'' as the evidence for the neutral forward-looking claim of the convenience store.
We also notice that models generate ``2019/20 estimated EPS is 9.69/10.71'' to the company, which EPS is always in the range of 1 to 3 (Numeral Hallucination).
The above case studies raise the second open issue: how to evaluate the fine-grained information in forward-looking claims? Additionally, because nobody knows about the future, does it mean models are wrong when the generated forward-looking claim is different from the ground truth?
For example, the predictions on EPS of the Ground Truth and Pegasus-News of N2A-2 in Table~\ref{tab:Cases Studies of News2Argument} are very close. 
How can we evaluate the numerals information in forward-looking claims if the true answer is not available at this time?
In light of these issues, we suggest future studies design evaluation methods from plausibility, suitableness, and likeliness aspects instead of only comparing models' results with human-written arguments.

Although the generated results of the forward-looking claim generation task are impressive with recent representative models, we find some worth discussing open questions based on the empirical results of our pilot explorations as follows: 

\vspace{1mm}

\noindent \textbf{(OQ1)}: How to avoid Past Event Hallucination and Causal Hallucination when generating forward-looking claims?

\vspace{1mm}

\noindent \textbf{(OQ2)}: How to evaluate the generated numeral information and avoid Numeral Hallucination when generating forward-looking claims?

\vspace{1mm}

\noindent \textbf{(OQ3)}: How to assess the trustworthiness and the forecasting skills of forward-looking claims when the truth is yet available?

\subsection{Scenario Planning}
\label{sec:Scenario Planning}
Scenario planning~\cite{kahn1967next,amer2013review}, which aims to create long-term strategic plans, necessitates extensive expert brainstorming and discussion.
Performing scenario planning for the future of public transportation and shared mobility during COVID-19 recovery, for example, requires up to 18 experts. Within three months, 23 hours were spent in four discussion phases~\cite{shaheen2021future}.
During the discussions, experts must imagine possible future scenarios and propose a tailored plan for each scenario. 
As shown in the News2Argument example in Table~\ref{fig:Example}, experts may pay less attention to what has already occurred and instead focus on plausible future scenarios and make forward-looking claims based on these scenarios rather than events that have already occurred. 
Although \citet{hashimoto-etal-2014-toward} quote scenario planning when dealing with future event prediction, we argue that it still has some differences between predicting known possible events and planning plausible but unprecedented scenarios. 
The application proposed in the previous work~\cite{hashimoto-etal-2014-toward} is similar to the forward-looking claim level instead of scenario planning.
The scenario planning for COVID-19 recovery~\cite{shaheen2021future} is one of the good cases, in which there are few pieces of evidence and previous discussions for models to learn from.

By observing the generated results of all proposed tasks, we notice that models can learn how to generate forward-looking claims but have not learned how to perform scenario planning yet. 
In most cases, models summarize the given input, then make the forward-looking claims with the sentences starting with ``we expect'' (Evidence2Claim and Chart2Argument) or the sentences with bullish/neutral/bearish opinions (News2Argument). 
Taking the N2A-1 in Table~\ref{tab:Cases Studies of News2Argument} as an example, an analyst (ground truth) first performs scenario planning for the next quarter's revenue with both evidence and forecasting, and then makes forward-looking claims based on the scenarios instead of the given news.
Models, on the other hand, only generate a summary of past events and then make forward-looking claims based on the past events. 
This raises an important open question: how to let models learn to imagine the future?

Although scenario planning has been used in many real-world applications for a long-time, such as business, policy, investment, electronic circuits, drug control, ecosystem, health care, forecasting petroleum consumption, and so on~\cite{amer2013review}, we think that the discussion on how NLP can assist experts in scenario planning is still in its early stages. We hope that the pilot results of this work and our discussions will inspire some researchers to join to accelerate the development of this direction. 
To sum up, we distinguish scenario planning from forward-looking claim generation (forecasting) in this section, note that current representative models are not yet capable of performing scenario planning. 
Thus, the following two open questions remain for future work:

\vspace{1mm}

\noindent \textbf{(OQ4)}: How to let models learn to generate plausible scenarios? 

\vspace{1mm}

\noindent \textbf{(OQ5)}: To what extent NLP methods can help experts in scenario planning?

\section{Limitation}
The limitations of this paper are three-fold as follows:
First, although we experimented with the data written in English and Chinese, we did not experiment with multilingual data on the same source. 
That limits us to discuss the difference between multilingual earnings conference calls, blogs, and professional reports.
We believe that this is an interesting direction for future work.
    
Second, although we tried our best to collect as much data as possible, the sizes of the datasets for Chart2Argument and News2Argument are still relatively small. 
It is a natural limitation of this research direction because it is costly to collect datasets written by professionals. 
Even though we have already collected all news-argument for more than one year, it is still only about two thousand instances. 
It is intuitive that analysts will not write down the comments on all news. 
Thus, we suggest that future work can explore the few-shot setting, which is closer to real-world situations, with the proposed dataset.
    
Third, because it is a bit hard to find well-formed documents with the long-term records for the Chart2Argument task, we can only find one kind of proper source, i.e., Elliott Wave Forecast Blog. 
That makes us can only explore the models' ability based on one technical analysis method, Elliott Wave Theory. 
Since there are many kinds of technical analysis methods, future work can explore the proposed task from different aspects with various technical analysis methods.

\section{Conclusion}
This paper introduces three innovative tasks centered on the generation of arguments in finance, accompanied by our preliminary findings. 
To enhance performance, we've crafted a method anchored by key terms and pinpoint two significant hurdles when generating arguments within the financial application context. 
Through our in-depth analysis, we present insights enriched by case studies. Furthermore, we spotlight two primary trajectories for subsequent research.
We aspire for this study to serve as a foundational reference in the discourse on argument generation in finance.

\section*{Acknowledgements}
This paper is based on results obtained from a project JPNP20006, commissioned by the New Energy and Industrial Technology Development
Organization (NEDO). The work of Chung-Chi Chen was supported in part by JSPS KAKENHI Grant Number 23K16956. 

\bibliography{custom}

\begin{thebibliography}{47}
\expandafter\ifx\csname natexlab\endcsname\relax\def\natexlab#1{#1}\fi

\bibitem[{Alhamzeh et~al.(2022)Alhamzeh, Lacin, and Egyed-Zsigmond}]{alhamzeh2022passau21}
Alaa Alhamzeh, M~K{\"u}rsad Lacin, and El{\H{o}}d Egyed-Zsigmond. 2022.
\newblock Passau21 at the ntcir-16 finnum-3 task: Prediction of numerical claims in the earnings calls with transfer learning.
\newblock In \emph{Proceedings of the 16th NTCIR Conference on Evaluation of Information Access Technologies}.

\bibitem[{Alshomary et~al.(2021)Alshomary, Chen, Gurcke, and Wachsmuth}]{alshomary-etal-2021-belief}
Milad Alshomary, Wei-Fan Chen, Timon Gurcke, and Henning Wachsmuth. 2021.
\newblock \href {https://doi.org/10.18653/v1/2021.eacl-main.17} {Belief-based generation of argumentative claims}.
\newblock In \emph{Proceedings of the 16th Conference of the European Chapter of the Association for Computational Linguistics: Main Volume}, pages 224--233, Online. Association for Computational Linguistics.

\bibitem[{Alshomary et~al.(2022)Alshomary, El~Baff, Gurcke, and Wachsmuth}]{alshomary-etal-2022-moral}
Milad Alshomary, Roxanne El~Baff, Timon Gurcke, and Henning Wachsmuth. 2022.
\newblock \href {https://doi.org/10.18653/v1/2022.acl-long.601} {The moral debater: A study on the computational generation of morally framed arguments}.
\newblock In \emph{Proceedings of the 60th Annual Meeting of the Association for Computational Linguistics (Volume 1: Long Papers)}, pages 8782--8797, Dublin, Ireland. Association for Computational Linguistics.

\bibitem[{Amer et~al.(2013)Amer, Daim, and Jetter}]{amer2013review}
Muhammad Amer, Tugrul~U Daim, and Antonie Jetter. 2013.
\newblock A review of scenario planning.
\newblock \emph{Futures}, 46:23--40.

\bibitem[{Atanasova et~al.(2020)Atanasova, Wright, and Augenstein}]{atanasova-etal-2020-generating}
Pepa Atanasova, Dustin Wright, and Isabelle Augenstein. 2020.
\newblock \href {https://doi.org/10.18653/v1/2020.emnlp-main.256} {Generating label cohesive and well-formed adversarial claims}.
\newblock In \emph{Proceedings of the 2020 Conference on Empirical Methods in Natural Language Processing (EMNLP)}, pages 3168--3177, Online. Association for Computational Linguistics.

\bibitem[{Bao et~al.(2021)Bao, Dong, and Wei}]{bao2021beit}
Hangbo Bao, Li~Dong, and Furu Wei. 2021.
\newblock {BE}i{T}: Bert pre-training of image transformers.
\newblock \emph{arXiv preprint arXiv:2106.08254}.

\bibitem[{Bozanic et~al.(2018)Bozanic, Roulstone, and Van~Buskirk}]{bozanic2018management}
Zahn Bozanic, Darren~T Roulstone, and Andrew Van~Buskirk. 2018.
\newblock Management earnings forecasts and other forward-looking statements.
\newblock \emph{Journal of Accounting and Economics}, 65(1):1--20.

\bibitem[{Cazier et~al.(2020)Cazier, Merkley, and Treu}]{cazier2020firms}
Richard~A Cazier, Kenneth~J Merkley, and John~S Treu. 2020.
\newblock When are firms sued for qualitative disclosures? implications of the safe harbor for forward-looking statements.
\newblock \emph{The Accounting Review}, 95(1):31--55.

\bibitem[{Chen et~al.(2021)Chen, Huang, and Chen}]{chen2021opinion}
Chung-Chi Chen, Hen-Hsen Huang, and Hsin-Hsi Chen. 2021.
\newblock \emph{From Opinion Mining to Financial Argument Mining}.
\newblock Springer Nature.

\bibitem[{Devlin et~al.(2019)Devlin, Chang, Lee, and Toutanova}]{devlin-etal-2019-bert}
Jacob Devlin, Ming-Wei Chang, Kenton Lee, and Kristina Toutanova. 2019.
\newblock \href {https://doi.org/10.18653/v1/N19-1423} {{BERT}: Pre-training of deep bidirectional transformers for language understanding}.
\newblock In \emph{Proceedings of the 2019 Conference of the North {A}merican Chapter of the Association for Computational Linguistics: Human Language Technologies, Volume 1 (Long and Short Papers)}, pages 4171--4186, Minneapolis, Minnesota. Association for Computational Linguistics.

\bibitem[{Dosovitskiy et~al.(2021)Dosovitskiy, Beyer, Kolesnikov, Weissenborn, Zhai, Unterthiner, Dehghani, Minderer, Heigold, Gelly et~al.}]{dosovitskiy2021image}
Alexey Dosovitskiy, Lucas Beyer, Alexander Kolesnikov, Dirk Weissenborn, Xiaohua Zhai, Thomas Unterthiner, Mostafa Dehghani, Matthias Minderer, Georg Heigold, Sylvain Gelly, et~al. 2021.
\newblock An image is worth 16x16 words: Transformers for image recognition at scale.
\newblock In \emph{International Conference on Learning Representations}.

\bibitem[{El~Baff et~al.(2019)El~Baff, Wachsmuth, Al~Khatib, Stede, and Stein}]{el-baff-etal-2019-computational}
Roxanne El~Baff, Henning Wachsmuth, Khalid Al~Khatib, Manfred Stede, and Benno Stein. 2019.
\newblock \href {https://doi.org/10.18653/v1/W19-8607} {Computational argumentation synthesis as a language modeling task}.
\newblock In \emph{Proceedings of the 12th International Conference on Natural Language Generation}, pages 54--64, Tokyo, Japan. Association for Computational Linguistics.

\bibitem[{Gretz et~al.(2020)Gretz, Bilu, Cohen-Karlik, and Slonim}]{gretz-etal-2020-workweek}
Shai Gretz, Yonatan Bilu, Edo Cohen-Karlik, and Noam Slonim. 2020.
\newblock \href {https://doi.org/10.18653/v1/2020.findings-emnlp.47} {The workweek is the best time to start a family {--} a study of {GPT}-2 based claim generation}.
\newblock In \emph{Findings of the Association for Computational Linguistics: EMNLP 2020}, pages 528--544, Online. Association for Computational Linguistics.

\bibitem[{Hashimoto et~al.(2014)Hashimoto, Torisawa, Kloetzer, Sano, Varga, Oh, and Kidawara}]{hashimoto-etal-2014-toward}
Chikara Hashimoto, Kentaro Torisawa, Julien Kloetzer, Motoki Sano, Istv{\'a}n Varga, Jong-Hoon Oh, and Yutaka Kidawara. 2014.
\newblock \href {https://doi.org/10.3115/v1/P14-1093} {Toward future scenario generation: Extracting event causality exploiting semantic relation, context, and association features}.
\newblock In \emph{Proceedings of the 52nd Annual Meeting of the Association for Computational Linguistics (Volume 1: Long Papers)}, pages 987--997, Baltimore, Maryland. Association for Computational Linguistics.

\bibitem[{Hidey and McKeown(2019)}]{hidey-mckeown-2019-fixed}
Christopher Hidey and Kathy McKeown. 2019.
\newblock \href {https://doi.org/10.18653/v1/N19-1174} {Fixed that for you: Generating contrastive claims with semantic edits}.
\newblock In \emph{Proceedings of the 2019 Conference of the North {A}merican Chapter of the Association for Computational Linguistics: Human Language Technologies, Volume 1 (Long and Short Papers)}, pages 1756--1767, Minneapolis, Minnesota. Association for Computational Linguistics.

\bibitem[{Holtermann et~al.(2022)Holtermann, Lauscher, and Ponzetto}]{holtermann-etal-2022-fair}
Carolin Holtermann, Anne Lauscher, and Simone Ponzetto. 2022.
\newblock \href {https://doi.org/10.18653/v1/2022.acl-long.541} {Fair and argumentative language modeling for computational argumentation}.
\newblock In \emph{Proceedings of the 60th Annual Meeting of the Association for Computational Linguistics (Volume 1: Long Papers)}, pages 7841--7861, Dublin, Ireland. Association for Computational Linguistics.

\bibitem[{Hu et~al.(2022)Hu, Chan, Liu, Xiao, Wu, and Huang}]{hu-etal-2022-planet}
Zhe Hu, Hou~Pong Chan, Jiachen Liu, Xinyan Xiao, Hua Wu, and Lifu Huang. 2022.
\newblock \href {https://doi.org/10.18653/v1/2022.acl-long.163} {{PLANET}: Dynamic content planning in autoregressive transformers for long-form text generation}.
\newblock In \emph{Proceedings of the 60th Annual Meeting of the Association for Computational Linguistics (Volume 1: Long Papers)}, pages 2288--2305, Dublin, Ireland. Association for Computational Linguistics.

\bibitem[{Hua et~al.(2019)Hua, Hu, and Wang}]{hua-etal-2019-argument-generation}
Xinyu Hua, Zhe Hu, and Lu~Wang. 2019.
\newblock \href {https://doi.org/10.18653/v1/P19-1255} {Argument generation with retrieval, planning, and realization}.
\newblock In \emph{Proceedings of the 57th Annual Meeting of the Association for Computational Linguistics}, pages 2661--2672, Florence, Italy. Association for Computational Linguistics.

\bibitem[{Hua and Wang(2018)}]{hua-wang-2018-neural}
Xinyu Hua and Lu~Wang. 2018.
\newblock \href {https://doi.org/10.18653/v1/P18-1021} {Neural argument generation augmented with externally retrieved evidence}.
\newblock In \emph{Proceedings of the 56th Annual Meeting of the Association for Computational Linguistics (Volume 1: Long Papers)}, pages 219--230, Melbourne, Australia. Association for Computational Linguistics.

\bibitem[{Hua and Wang(2020)}]{hua-wang-2020-pair}
Xinyu Hua and Lu~Wang. 2020.
\newblock \href {https://doi.org/10.18653/v1/2020.emnlp-main.57} {{PAIR}: Planning and iterative refinement in pre-trained transformers for long text generation}.
\newblock In \emph{Proceedings of the 2020 Conference on Empirical Methods in Natural Language Processing (EMNLP)}, pages 781--793, Online. Association for Computational Linguistics.

\bibitem[{Kahn and Wiener(1967)}]{kahn1967next}
Herman Kahn and Anthony~J Wiener. 1967.
\newblock The next thirty-three years: A framework for speculation.
\newblock \emph{Daedalus}, pages 705--732.

\bibitem[{Lewis et~al.(2020)Lewis, Liu, Goyal, Ghazvininejad, Mohamed, Levy, Stoyanov, and Zettlemoyer}]{lewis2020bart}
Mike Lewis, Yinhan Liu, Naman Goyal, Marjan Ghazvininejad, Abdelrahman Mohamed, Omer Levy, Veselin Stoyanov, and Luke Zettlemoyer. 2020.
\newblock {BART}: Denoising sequence-to-sequence pre-training for natural language generation, translation, and comprehension.
\newblock In \emph{Proceedings of the 58th Annual Meeting of the Association for Computational Linguistics}, pages 7871--7880.

\bibitem[{Li et~al.(2022)Li, Zhu, Thomas, Rudzicz, and Xu}]{li-etal-2022-neural}
Bai Li, Zining Zhu, Guillaume Thomas, Frank Rudzicz, and Yang Xu. 2022.
\newblock \href {https://doi.org/10.18653/v1/2022.acl-long.512} {Neural reality of argument structure constructions}.
\newblock In \emph{Proceedings of the 60th Annual Meeting of the Association for Computational Linguistics (Volume 1: Long Papers)}, pages 7410--7423, Dublin, Ireland. Association for Computational Linguistics.

\bibitem[{Li(2010)}]{li2010information}
Feng Li. 2010.
\newblock The information content of forward-looking statements in corporate filings—a na{\"\i}ve bayesian machine learning approach.
\newblock \emph{Journal of Accounting Research}, 48(5):1049--1102.

\bibitem[{Li et~al.(2021)Li, Lv, Cui, Lu, Florencio, Zhang, Li, and Wei}]{li2021trocr}
Minghao Li, Tengchao Lv, Lei Cui, Yijuan Lu, Dinei Florencio, Cha Zhang, Zhoujun Li, and Furu Wei. 2021.
\newblock Trocr: Transformer-based optical character recognition with pre-trained models.
\newblock \emph{arXiv preprint arXiv:2109.10282}.

\bibitem[{Lin(2004)}]{lin-2004-rouge}
Chin-Yew Lin. 2004.
\newblock \href {https://www.aclweb.org/anthology/W04-1013} {{ROUGE}: A package for automatic evaluation of summaries}.
\newblock In \emph{Text Summarization Branches Out}, pages 74--81, Barcelona, Spain. Association for Computational Linguistics.

\bibitem[{Liu et~al.(2019)Liu, Ott, Goyal, Du, Joshi, Chen, Levy, Lewis, Zettlemoyer, and Stoyanov}]{liu2019roberta}
Yinhan Liu, Myle Ott, Naman Goyal, Jingfei Du, Mandar Joshi, Danqi Chen, Omer Levy, Mike Lewis, Luke Zettlemoyer, and Veselin Stoyanov. 2019.
\newblock Roberta: A robustly optimized bert pretraining approach.
\newblock \emph{arXiv preprint arXiv:1907.11692}.

\bibitem[{Liu et~al.(2021)Liu, Lin, Cao, Hu, Wei, Zhang, Lin, and Guo}]{liu2021swin}
Ze~Liu, Yutong Lin, Yue Cao, Han Hu, Yixuan Wei, Zheng Zhang, Stephen Lin, and Baining Guo. 2021.
\newblock Swin transformer: Hierarchical vision transformer using shifted windows.
\newblock In \emph{Proceedings of the IEEE/CVF International Conference on Computer Vision}, pages 10012--10022.

\bibitem[{Mori et~al.(2020)Mori, Hirakawa, Yamashita, and Fujiyoshi}]{mori2020image}
Yuki Mori, Tsubasa Hirakawa, Takayoshi Yamashita, and Hironobu Fujiyoshi. 2020.
\newblock Image captioning in near-future from vehicle camera images and motion information.
\newblock \emph{IEICE Technical Report; IEICE Tech. Rep.}, 120(154):13--18.

\bibitem[{Muslu et~al.(2015)Muslu, Radhakrishnan, Subramanyam, and Lim}]{muslu2015forward}
Volkan Muslu, Suresh Radhakrishnan, KR~Subramanyam, and Dongkuk Lim. 2015.
\newblock Forward-looking {MD}\&{A} disclosures and the information environment.
\newblock \emph{Management Science}, 61(5):931--948.

\bibitem[{Narayan et~al.(2022)Narayan, Sim{\~o}es, Zhao, Maynez, Das, Collins, and Lapata}]{narayan-etal-2022-well}
Shashi Narayan, Gon{\c{c}}alo Sim{\~o}es, Yao Zhao, Joshua Maynez, Dipanjan Das, Michael Collins, and Mirella Lapata. 2022.
\newblock \href {https://doi.org/10.18653/v1/2022.acl-long.94} {A well-composed text is half done! composition sampling for diverse conditional generation}.
\newblock In \emph{Proceedings of the 60th Annual Meeting of the Association for Computational Linguistics (Volume 1: Long Papers)}, pages 1319--1339, Dublin, Ireland. Association for Computational Linguistics.

\bibitem[{Narayan et~al.(2021)Narayan, Zhao, Maynez, Sim{\~o}es, Nikolaev, and McDonald}]{narayan2021planning}
Shashi Narayan, Yao Zhao, Joshua Maynez, Gon{\c{c}}alo Sim{\~o}es, Vitaly Nikolaev, and Ryan McDonald. 2021.
\newblock Planning with learned entity prompts for abstractive summarization.
\newblock \emph{Transactions of the Association for Computational Linguistics}, 9:1475--1492.

\bibitem[{Pan et~al.(2021)Pan, Chen, Xiong, Kan, and Wang}]{pan-etal-2021-zero}
Liangming Pan, Wenhu Chen, Wenhan Xiong, Min-Yen Kan, and William~Yang Wang. 2021.
\newblock \href {https://doi.org/10.18653/v1/2021.acl-short.61} {Zero-shot fact verification by claim generation}.
\newblock In \emph{Proceedings of the 59th Annual Meeting of the Association for Computational Linguistics and the 11th International Joint Conference on Natural Language Processing (Volume 2: Short Papers)}, pages 476--483, Online. Association for Computational Linguistics.

\bibitem[{Raffel et~al.(2020)Raffel, Shazeer, Roberts, Lee, Narang, Matena, Zhou, Li, and Liu}]{raffel2020exploring}
Colin Raffel, Noam Shazeer, Adam Roberts, Katherine Lee, Sharan Narang, Michael Matena, Yanqi Zhou, Wei Li, and Peter~J Liu. 2020.
\newblock Exploring the limits of transfer learning with a unified text-to-text transformer.
\newblock \emph{Journal of Machine Learning Research}, 21:1--67.

\bibitem[{Saint-Dizier(2016)}]{saint-dizier-2016-challenges}
Patrick Saint-Dizier. 2016.
\newblock \href {https://doi.org/10.18653/v1/W16-6613} {Challenges of argument mining: Generating an argument synthesis based on the qualia structure}.
\newblock In \emph{Proceedings of the 9th International Natural Language Generation conference}, pages 79--83, Edinburgh, UK. Association for Computational Linguistics.

\bibitem[{Sato et~al.(2015)Sato, Yanai, Miyoshi, Yanase, Iwayama, Sun, and Niwa}]{sato-etal-2015-end}
Misa Sato, Kohsuke Yanai, Toshinori Miyoshi, Toshihiko Yanase, Makoto Iwayama, Qinghua Sun, and Yoshiki Niwa. 2015.
\newblock \href {https://doi.org/10.3115/v1/P15-4019} {End-to-end argument generation system in debating}.
\newblock In \emph{Proceedings of {ACL}-{IJCNLP} 2015 System Demonstrations}, pages 109--114, Beijing, China. Association for Computational Linguistics and The Asian Federation of Natural Language Processing.

\bibitem[{Schiller et~al.(2021)Schiller, Daxenberger, and Gurevych}]{schiller-etal-2021-aspect}
Benjamin Schiller, Johannes Daxenberger, and Iryna Gurevych. 2021.
\newblock \href {https://doi.org/10.18653/v1/2021.naacl-main.34} {Aspect-controlled neural argument generation}.
\newblock In \emph{Proceedings of the 2021 Conference of the North American Chapter of the Association for Computational Linguistics: Human Language Technologies}, pages 380--396, Online. Association for Computational Linguistics.

\bibitem[{Shaheen and Wong(2021)}]{shaheen2021future}
Susan Shaheen and Stephen Wong. 2021.
\newblock Future of public transit and shared mobility: Scenario planning for covid-19 recovery.

\bibitem[{Touvron et~al.(2021)Touvron, Cord, Douze, Massa, Sablayrolles, and J{\'e}gou}]{touvron2021training}
Hugo Touvron, Matthieu Cord, Matthijs Douze, Francisco Massa, Alexandre Sablayrolles, and Herv{\'e} J{\'e}gou. 2021.
\newblock Training data-efficient image transformers \& distillation through attention.
\newblock In \emph{International Conference on Machine Learning}, pages 10347--10357. PMLR.

\bibitem[{Wachsmuth et~al.(2018)Wachsmuth, Stede, El~Baff, Al-Khatib, Skeppstedt, and Stein}]{wachsmuth-etal-2018-argumentation}
Henning Wachsmuth, Manfred Stede, Roxanne El~Baff, Khalid Al-Khatib, Maria Skeppstedt, and Benno Stein. 2018.
\newblock \href {https://aclanthology.org/C18-1318} {Argumentation synthesis following rhetorical strategies}.
\newblock In \emph{Proceedings of the 27th International Conference on Computational Linguistics}, pages 3753--3765, Santa Fe, New Mexico, USA. Association for Computational Linguistics.

\bibitem[{Wambsganss and Niklaus(2022)}]{wambsganss-niklaus-2022-modeling}
Thiemo Wambsganss and Christina Niklaus. 2022.
\newblock \href {https://doi.org/10.18653/v1/2022.acl-long.599} {Modeling persuasive discourse to adaptively support students{'} argumentative writing}.
\newblock In \emph{Proceedings of the 60th Annual Meeting of the Association for Computational Linguistics (Volume 1: Long Papers)}, pages 8748--8760, Dublin, Ireland. Association for Computational Linguistics.

\bibitem[{Wang and Hussainey(2013)}]{wang2013voluntary}
Mingzhu Wang and Khaled Hussainey. 2013.
\newblock Voluntary forward-looking statements driven by corporate governance and their value relevance.
\newblock \emph{Journal of accounting and public policy}, 32(3):26--49.

\bibitem[{Wright et~al.(2022)Wright, Wadden, Lo, Kuehl, Cohan, Augenstein, and Wang}]{wright-etal-2022-generating}
Dustin Wright, David Wadden, Kyle Lo, Bailey Kuehl, Arman Cohan, Isabelle Augenstein, and Lucy Wang. 2022.
\newblock \href {https://doi.org/10.18653/v1/2022.acl-long.175} {Generating scientific claims for zero-shot scientific fact checking}.
\newblock In \emph{Proceedings of the 60th Annual Meeting of the Association for Computational Linguistics (Volume 1: Long Papers)}, pages 2448--2460, Dublin, Ireland. Association for Computational Linguistics.

\bibitem[{Yasunaga et~al.(2022)Yasunaga, Leskovec, and Liang}]{yasunaga-etal-2022-linkbert}
Michihiro Yasunaga, Jure Leskovec, and Percy Liang. 2022.
\newblock \href {https://aclanthology.org/2022.acl-long.551} {{L}ink{BERT}: Pretraining language models with document links}.
\newblock In \emph{Proceedings of the 60th Annual Meeting of the Association for Computational Linguistics (Volume 1: Long Papers)}, pages 8003--8016, Dublin, Ireland. Association for Computational Linguistics.

\bibitem[{Zhang et~al.(2020)Zhang, Zhao, Saleh, and Liu}]{zhang2020pegasus}
Jingqing Zhang, Yao Zhao, Mohammad Saleh, and Peter Liu. 2020.
\newblock Pegasus: Pre-training with extracted gap-sentences for abstractive summarization.
\newblock In \emph{International Conference on Machine Learning}, pages 11328--11339. PMLR.

\bibitem[{Zhang et~al.(2019)Zhang, Kishore, Wu, Weinberger, and Artzi}]{zhang2019bertscore}
Tianyi Zhang, Varsha Kishore, Felix Wu, Kilian~Q Weinberger, and Yoav Artzi. 2019.
\newblock {BERTS}core: Evaluating text generation with {BERT}.
\newblock In \emph{International Conference on Learning Representations}.

\bibitem[{Zhao et~al.(2019)Zhao, Chen, Zhang, Zhao, Liu, Lu, Chen, Deng, Ju, and Du}]{zhao2019uer}
Zhe Zhao, Hui Chen, Jinbin Zhang, Xin Zhao, Tao Liu, Wei Lu, Xi~Chen, Haotang Deng, Qi~Ju, and Xiaoyong Du. 2019.
\newblock Uer: An open-source toolkit for pre-training models.
\newblock page 241.

\end{thebibliography}

\end{document}